\documentclass[10pt,journal,compsoc]{IEEEtran}

\ifCLASSOPTIONcompsoc

\usepackage[nocompress]{cite}
\else
\usepackage{cite}
\fi
\usepackage{color}
\usepackage{times}
\usepackage{epsfig}
\usepackage{graphicx}
\usepackage{amsmath}
\usepackage{amssymb}
\usepackage{subfigure}
\usepackage{multirow}
\usepackage{caption}
\usepackage{float}
\usepackage{enumitem}   
\usepackage{makecell}
\usepackage{epstopdf}

\newcommand{\etal}{\textit{et al.}}

\newtheorem{lemma}{Lemma}

\ifCLASSINFOpdf
\else
\fi
\begin{document}
	%
	
	\title{A Global Alignment Kernel based Approach for Group-level Happiness Intensity Estimation}
	
	\author{Xiaohua~Huang,
		Abhinav Dhall, Roland Goecke, \textit{Member, IEEE},
		Matti Pietik\"{a}inen, \textit{Fellow, IEEE},
		and
		Guoying~Zhao, \textit{Senior Member, IEEE}
		\IEEEcompsocitemizethanks{\IEEEcompsocthanksitem X. Huang, M. Pietik\"{a}inen and G. Zhao are with the Center for Machine Vision and Signal Analysis, University of Oulu, Finland. \protect\\
			E-mail: xiaohua.huang@oulu.fi, mkp@ee.oulu.fi, gyzhao@ee.oulu.fi
			
			\IEEEcompsocthanksitem A. Dhall is with the Department of Computer Science and Engineering, Indian
			Institute of Technology, Ropar, India. \protect\\
			E-mail: abhinav@iitrpr.ac.in
			
			\IEEEcompsocthanksitem R. Goecke is with Human-Centred Technology Research Centre, University
			of Canberra, Australia.
			\protect\\
			E-mail: Roland.Goecke@canberra.edu.au
			
		}
		\thanks{}}
	
	\markboth{Journal of \LaTeX\ Class Files,~Vol.~14, No.~8, August~2017}%
	{Shell \MakeLowercase{\textit{et al.}}: Bare Demo of IEEEtran.cls for Computer Society Journals}
	
	\IEEEtitleabstractindextext{%
		\begin{abstract}
			With the progress in automatic human behavior understanding, analysing the perceived affect of multiple people has been recieved interest in affective computing community. Unlike conventional facial expression analysis, this paper primarily focuses on analysing the behaviour of multiple people in an image. The proposed method is based on support vector regression with the combined global alignment kernels (GAKs) to estimate the happiness intensity of a group of people. We first exploit Riesz-based volume local binary pattern (RVLBP) and deep convolutional neural network (CNN) based features for characterizing facial images. Furthermore, we propose to use the GAK for RVLBP and deep CNN features, respectively for explicitly measuring the similarity of two group-level images. Specifically, we exploit the global weight sort scheme to sort the face images from group-level image according to their spatial weights, making an efficient data structure to GAK. Lastly, we propose Multiple kernel learning based on three combination strategies for combining two respective GAKs based on RVLBP and deep CNN features, such that enhancing the discriminative ability of each GAK. Intensive experiments are performed on the challenging group-level happiness intensity database, namely HAPPEI. Our experimental results demonstrate that the proposed approach achieves promising performance for group happiness intensity analysis, when compared with the recent state-of-the-art methods.
		\end{abstract}
		
		}

	\maketitle
	
	\IEEEdisplaynontitleabstractindextext
	
	\IEEEpeerreviewmaketitle
	
	\section{Introduction}
	\label{sec:intro}
	
\begin{figure*}[t!]
		\centering
		\subfigure[]{
			\label{fig:fig1a}
			\includegraphics[width=0.25\linewidth]{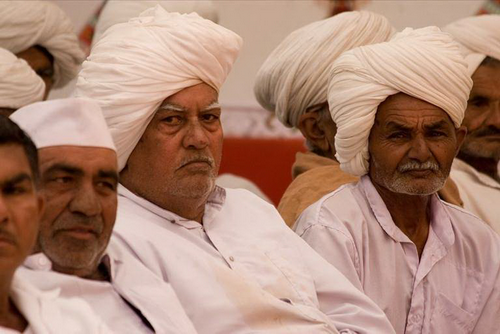}}
		\subfigure[]{
			\label{fig:fig1b}
			\includegraphics[width=0.25\linewidth]{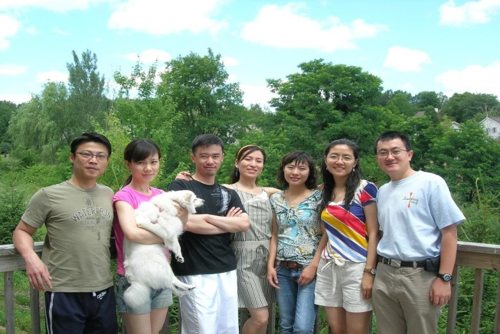}}
		\subfigure[]{
			\label{fig:fig1c}
			\includegraphics[width=0.25\linewidth]{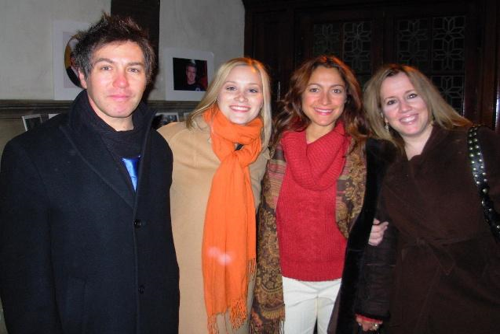}}
		\subfigure[]{
			\label{fig:fig1d}
			\includegraphics[width=0.25\linewidth]{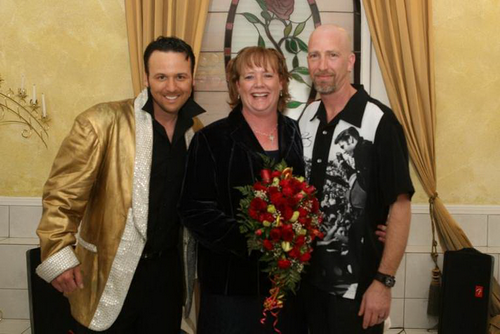}}
		\subfigure[]{
			\label{fig:fig1e}
			\includegraphics[width=0.25\linewidth]{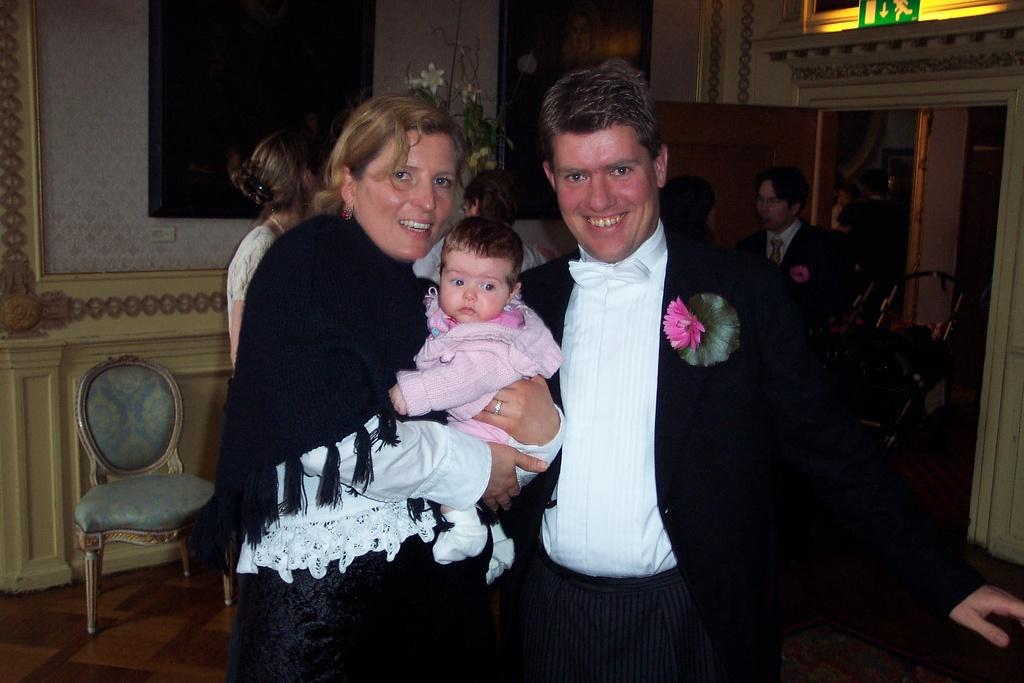}}
		\subfigure[]{
			\label{fig:fig1f}
			\includegraphics[width=0.25\linewidth]{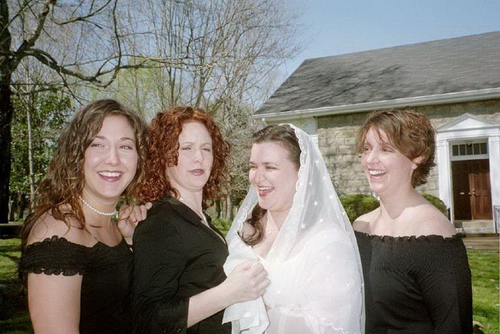}}
		\caption{Six sample images containing groups in social events. Happiness intensity scores of (a), (b), (c), (d), (e) and (f) are 0, 1, 2, 3, 4, and 5,  respectively. Source: HAPPEI database.}
		\label{fig:fig1}
	\end{figure*}

	With the credible progress in social media, millions of images are made available on the Internet through social networks such as Facebook. The large-scale data enables us to analyse the human behaviour during social events (for example Figure~\ref{fig:fig1}). Recently, several researches have studied group-level happiness intensity estimation. Specifically, they have concentrated on analysing the affect exhibited by multiple people in an image. Group-level happiness intensity estimation has various benefits to computer vision and multimedia~\cite{Dhall2015}.
	
	According to the social psychology findings~\cite{Kelly2001,Barsade1998}, the methods for theorizing group-level emotion can be commonly conceptualized into the bottom-up and top-down categories. The bottom-up schemes use the subject's attributes to infer group emotion. For example, in~\cite{Hernandez2012}, Hernandez et al. exploited the smile of each people as subject's attribute for inferring the emotion of crowd. The top-down methods consider external attributes, such as the affect of the scene and the position of people, to describing group members. For instance, Gallagher et al.~\cite{Gallagher2009} proposed contextual features based on the group structure for computing the age and gender of individuals. For group affective analysis, the bottom-up or top-down approach alone may miss some useful information exploiting from an image. For example, the bottom-up method may ignore the influence of scene to group-level emotion, while top-down approach does not consider the person's attributes such as intensity of facial expression. To alleviate that problem, a hybrid model was recently proposed by combining bottom-up and top-down components for group affective analysis. It can be grouped into group expression model (\textbf{GEM})~\cite{dhall2013finding,Dhall2015,Huang2015} and multi-modal framework~\cite{Dhall2015b,Mou2015,Cerekovic2016,Li2016,Sun2016}. GEM is defined as modelling the global and local social attributes based on a graph~\cite{Dhall2012}, while multi-modal framework aims to fuse multiple attribute features by using machine learning methods, e.g., multiple kernel learning. Specifically, multi-modal framework considers the method for aggregating specific social attribute from group-level image\footnote{Group-level image is defined as an image containing more than two faces as a group. For the purpose of simplicity, we use `image' to represent `group-level image'.}, such as face, into one compact feature vector. Although GEM methods and multi-modal framework explore useful information from images such as face, pose and scene, there is one difference between them: GEM uses temporal model, e.g., continuous conditional random fields, however multi-modal framework exploits feature aggregation methods and feature fusion approaches.

	According to our experimental experience on GEM~\cite{Huang2015} and multi-modal framework~\cite{Huang2016}, both methods are very complicate due to quite many adjust parameters. It is worth to consider whether there is a simply way for straightforwardly measuring the similarity of images, such that it can be flexibly embedded into any classifiers. This question leads to a relatively unexplored problem in group affective analysis: the alignment between faces in images. Therefore, unlike GEM and multi-modal framework, this paper focuses on an efficient, fast and simply way to measure the similarity of images. However, there are three challenges for the alignment between face in images: (1) the number of faces are not always consistent in two images, (2) the structure of groups in different images may vary and (3) it is non-trivial to compute alignment between different feature sets of multidimensional features. 
	
	Recently, time series kernel methods have been used to measure the similarity distance between two facial expression sequences~\cite{GA-Emo,GA-Emo2,CTW}. Additionally, time series kernel methods, especially global kernel alignment (\textbf{GAK}), have been demonstrated to be an efficient and efficiency scheme to simply measure the similarity of two image sequences. Inspired by the works~\cite{GA-Emo,GA-Emo2,CTW}, we first formulate the faces in an image like an image sequence, and then use GAK to measure the similarity of two images. For example, as illustrated in the Figure~\ref{fig:fig1b}, we may view this image as a face sequence containing 7 faces. Under this assumption, measuring pairwise image can be explicitly formulated as the alignment between two video sequences. However, two important issues need to be addressed in the alignment between images: (1) In practice, for an image, faces are sorted according to the order as returned by a face detector. The sequence of face detection further does not depend on the structure of a group. In~\cite{Dhall2015}, global weights of people are aggregated into GEM for emphasising the importance of people. Motivated by~\cite{Dhall2015}, person ranking can be used in this paper to formulate the faces in an image such like an an image sequence. It will make the consistent data structure to GAK. (2) On the other hand, face may suffer from problems caused by the challenging environment e.g., bad illumination. It is a promising way to explore multiple features for describing faces in images. For alignment, feature concatenation leads to the `curse of dimensionality' problem. It is still an open research question on how to efficiently combine multiple features on time series kernels. As we know, there are many way to combine multiple features, such as directly feature concatenation. Amongst fusion methods, multiple kernel learning (\textbf{MKL}) has commonly used in many fields, such as gene classification~\cite{Pavlidis2001, Ben2005}, and has been demonstrated to achieve promising performance when combining several kernels. Therefore, we attempt to exploit MKL method to combine two GAK kernels for group-level happiness intensity estimation.
	
	The \textbf{key-contributions} of this paper are as follows: (1) The temporal alignment method, especially GAK is the first time to be proposed for measuring similarity of two group-level images; (2) Global weight sorted scheme is developed to efficiently provide a consistent structure for GAK; (3) MKL based on summation, production and weighted summation strategies are proposed to aggregate two different features into GAK for inferring the perceived happiness intensity of a group of people in an image; (4) Comprehensive experiments on the `in the wild' database demonstrate the superiority of GAK over the state-of-the-art methods.
	
	The rest of the paper is organized as follows. Section~\ref{sec:relatedWork} describes the state-of-the-art methods in group-level happiness intensity estimation, time series kernel and two challenging databases. Section~\ref{sec:facialFeature} exploits two robust features for facial images in the challenging situations, as facial expression representation has an important role in characterizing the discriminative information among different facial expressions. Section~\ref{sec:methodlogy} discusses our method for exploring a simple but efficient way to measure the similarity of a group of faces in the images. Section~\ref{sec:experiment} presents the experiments results with empirical analysis. Section~\ref{sec:conclusion} concludes the paper.
	
	\section{Related work}
	\label{sec:relatedWork}
	\subsection{The state of the arts}
	
	Recently, several hybrid approaches have been developed to estimate happiness intensity of multiple people. They are categorized into two branches: GEM and multi-modal framework, 
	
	According to the literature review, GEM based approaches encode multiple faces in an image into a graph structure. The first one has been appeared in~\cite{dhall2013finding,Dhall2015}, where Dhall~\etal exploited three models, namely, average, weighted and Latent Dirichlet Allocation (LDA) based GEMs, for group-level happiness intensity estimation. Specifically, in their method, they used the effect of the event and surrounding of a group as the top-down component and the group members together with their attributes such as spontaneous expressions, clothes, age and gender as the bottom-up component. Huang~\etal~\cite{Huang2015} proposed one various GEM approach for increasing the performance of group-level happiness intensity estimation. In~\cite{Huang2015}, they referred global attributes such as the effect of neighbouring group members as the top-down component, and local attributes such as an individual's feature as the bottom-up component. However, the GEM based approaches are not efficient in computation and not stable due to graph construction and possible noise in the face descriptors. For example, in~\cite{Dhall2015}, the performance of GEM based on LDA is seriously effected by the choice of the facial expression descriptors and the change of the number of the visual words. For example, the large number of the visual words makes the feature very sparse while the small number losses the discriminative information. In~\cite{Huang2015}, the graph construction suffers from the false prediction of support vector regression. Additionally, GEM cannot directly measure the similarity of images by using statistical model such as the LDA. 
	
	Multi-modal framework is an alternative way in group affective analysis to combine bottom-up and top-down components of images. For example, in~\cite{Dhall2015b}, the facial action unit and face features are regarded as the bottom-up component, while scene feature is considered as the top-down component. Similar works have also appeared in~\cite{Cerekovic2016, Li2016,Sun2016}. Another interesting multi-modal work~\cite{Mou2015} combined face and body information to predict the valence and arousal of a group of people. It is seen that some works on multi-modal framework such as~\cite{Mou2015} prefer to set up the condition for group-level affect analysis, in which they experimented on specific groups based on the fixed number of faces and bodies. As well, the feature encoding methods proposed by~\cite{Dhall2015b} presented to use some clustering ways to encode feature of all faces into visual words. This intermediate stage may bring some errors at the classification stage. Additionally, these methods are seriously affected by the parameter design in clustering approaches.
	
	\subsection{Family of Time series kernels}
	
	Recently, a family of time series kernels based on dynamic programming was exploited for constructing kernel in speech, bi-informatics and text-processing. These time series kernels can resolve two critical issues: (1) the time series might be a variable length and (2) standard kernels for vectors cannot capture by constructing the local dependencies between neighbouring states of their time series, when measuring a varied length sequence. The time series kernel approach, such as Dynamic Time Warping (\textbf{DTW})~\cite{DTW1,DTW2}, has been investigated to align spatiotemporal facial expression~\cite{GA-Emo,GA-Emo2,CTW}, action recognition~\cite{Gaidon2011,Brun2016} and music retrieval~\cite{Deng2015}. However, such distances cannot be translated easily into positive definite (\textbf{PD}) kernels, which is an important requirement of kernel machines during the training phase. For addressing the PD problem of time series kernels, such as DTW, Cuturi~et.al proposed global alignment kernel (\textbf{GAK}) method with applications to speech recognition~\cite{GA} and handwritten recognition~\cite{FGA}. Recently, Support Vector Machine (\textbf{SVM}) based on GAK has been proposed to align the temporal information for dynamic facial expression recognition and has been shown its efficiency~\cite{GA-Emo,GA-Emo2}. 
	
	\begin{figure*}[t!]
		\centering
		\includegraphics[width=0.8\linewidth]{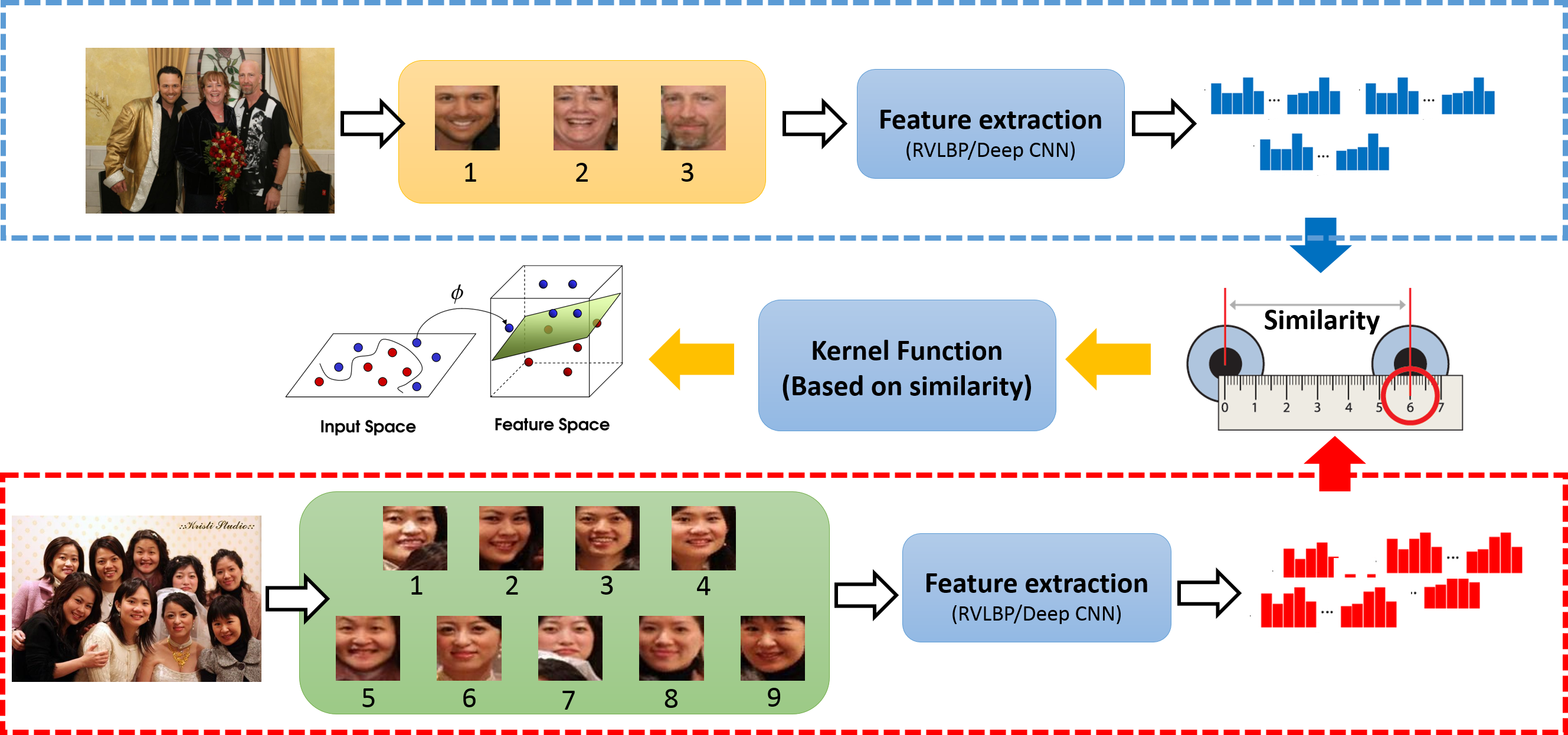}
		\caption{Illustration of the `Unequal Sample Size Problem' for measuring the similarity of two images. The number under face image represents order index of image obtained by face detection. The upper and bottom dotted-line blocks is the pipeline of feature extraction. The objective of addressing `Unequal Sample Size Problem' is to search proper similarity function and classifier, as shown in the middle pipeline.}
		\label{fig:fig2}
	\end{figure*}
	
	\section{Methodology}
	\label{sec:methodlogy}
	
	\section{Facial Expression Representation}
	\label{sec:facialFeature}
	
	To deal with the problems caused by the challenging environments, such as blurred faces and poor illumination, we use Riesz-based Volume Local Binary Pattern (\textbf{RVLBP}), which is proposed in~\cite{Huang2015} as the low-level feature, and deep onvolutional Neural Network (\textbf{CNN}) feature based on VGG-face, which is investigated in~\cite{Ng2015,Lopes2017}, as the high-level feature for time series kernel.
	
	\subsection{Problem Formulation}
	\label{sec:problemForm}
	
	Given an image $\mathbf{Y}$, we aim to estimate the happiness intensity using the faces based information. This task can be viewed as a regression analysis problem. There are lots of machine learning methods having been used to classify or estimate the data. Support vector regression (\textbf{SVR}) is one of the most successfully statistical pattern classifiers that utilize a kernel technique. Therefore, we propose to use the SVR to estimate the happiness intensity of a group of people in an image. For a given sample $\mathbf{Y}$, the basic form of SVR is expressed as,
	\begin{equation}
	\label{eqn:SVR}
	g(\mathbf{Y})=\sum_{i=1}^N\omega_il_i\Phi(\mathbf{X}_i)\cdot\Phi(\mathbf{Y})+b=\sum_{i=1}^{N}\omega_il_iK(\mathbf{X}_i,\mathbf{Y})+b,
	\end{equation}
	where $N$ is the number of images in the training set, $\Phi$ is a non-linear mapping function, `$\cdot$' denotes the inner product operator, $\mathbf{X}_i$, $l_i$ and $\omega_i$ are the $i$-th training sample, the corresponding class label, and its Lagrange multiplier, respectively, $K$ is a kernel function, and $b$ is a bias of SVR.
	
	As seen in Equation~\ref{eqn:SVR}, the kernel function $K$ has an important role in SVR, therefore we reconsider the definition of kernel function of pair-wise images in the task of the happiness intensity estimation. For group affective analysis, the number of faces is varied in each image. For example, in Figure~\ref{fig:fig2}, there are 3 and 9 faces existed in two upper-part and bottom-part images, respectively. Basically, we may conduct the subjects' number fixed strategy proposed in~\cite{Mou2015} that experimented on specific groups based on the fixed number of faces. Unfortunately, it may restrict the application of group affect analysis, especially the number of faces in an image was not in these specific groups. It leads to one critical question how to measure the similarity of two images for the kernel function $K$.  For simplicity, we name this case ``Unequal Sample Size Problem''. Numerous methods in machine learning can be used to resolve ``Unequal Sample Size Problem''. Specifically, a histogram can be calculated by using Bag-of-Word for representing $\mathbf{Y}$. For example, in~\cite{Dhall2015b}, Dhall et al. used Bag-of-Word to accumulate a histogram from multiple faces for a group image. Unfortunately, the feature obtained by using Bag-of-Word is very sparse. In~\cite{Huang2016}, they proposed an information aggregation method to encode the histograms of blocks of faces for representing an image. The approach gives advantage that the histogram is not sparse. However, their method suffers from parameter adjustment problem, i.e., there are quite many parameters need to be adequately adjusted, such as block number, PCA dimension and word size. Therefore, for group affective analysis, ``Unequal Sample Size Problem'' still becomes one critical and important problem when we want to simply but effectively measure the similarity of two images.
	
	Given two images, denoted as $\mathbf{X}_i$ and $\mathbf{X}_j$, there are $M_i$ and $M_j$ faces existing in $\mathbf{X}_i$ and $\mathbf{X}_j$, respectively. We extract their corresponding faces features denoted as $\{\mathbf{f}_{m}\}_{m=1}^{M_i}$ and $\{\mathbf{g}_{n}\}_{n=1}^{M_j}$ for the images $\mathbf{X}_i$ and $\mathbf{X}_j$, respectively. Thus, the similarity measurement among images is formulated as follows:
	\begin{equation}
	\label{eqn:problem}
	s(\mathbf{X}_i, \mathbf{X}_j)=f(\{\mathbf{f}_{1}, \mathbf{f}_{2}, \ldots, \mathbf{f}_{M_i}\}, \{\mathbf{g}_{1}, \mathbf{g}_{2}, \ldots, \mathbf{g}_{M_j}\}),
	\end{equation}
	where $s$ is the similarity between images, and $f$ means the similarity measurement function that attempts to calculate the implicit distance of two feature sets $\{\mathbf{f}_{1}, \mathbf{f}_{2}, \ldots, \mathbf{f}_{M_i}\}$ and $\{\mathbf{g}_{1}, \mathbf{g}_{2}, \ldots, \mathbf{g}_{M_j}\}$. In next section, we will discuss how to derive the similarity measurement function $f$ in Equation~\ref{eqn:problem} for measuring two images.
	
	\subsection{Global Alignment Kernels with global weight sort}
	\label{sec:PDK}
	
	As mentioned in the previous section, the number of faces in groups may vary from one image to another. Additionally, the spatial structures of faces in different images may vary. In this part, we propose `global weight', namely ``Global weight sort' for GAK, and describe our proposed method for group-level happiness intensity estimation.
	
	\subsubsection{Global weight sort of faces}
	\label{sec:GWS}
	
	Given an image $\mathbf{X}_i$, there are $M_i$ face detected, denoted as $\mathbf{x}_1,\ldots,\mathbf{x}_{M_i}$. A fully connected graph $\mathcal{G}=(\mathbf{I}, \mathbf{E})$ is constructed to map the global structure of faces in a group, where an edge $\mathbf{E}_{k,l}$ represents the link between $\mathbf{x}_k$ and $\mathbf{x}_l$. For obtaining $\mathcal{G}$, the minimal spanning tree algorithm~\cite{Prim1957} is implemented, providing the location and minimally connected neighbours of a face. Based on $\mathcal{G}$, we obtain the global weight of $\mathbf{x}_k$ using relative face size $S_{k}$ and relative distance $\delta_{k}$, where we present the same way to~\cite{Huang2015} to calculate $S_{k}$ and $\delta_{k}$.
	
	For $\mathbf{x}_k$, the size of face is taken as the distance between the locations of the left and right eyes, in which the distance is calculated by using $d_k=\parallel d_{L,k}-d_{R,k}\parallel$. The relative face size $S_{k}$ of $\mathbf{x}_k$ is given by $S_{k}=\frac{d_k}{\sum_{j=1}^{n}\frac{d_j}{n}}$, where $n$ is the number of neighbouring faces of $\mathbf{x}_k$.
	
	For relative face distance, based on the nose tip locations of all faces in an image, their centroid $c_g$ is computed by using $\frac{\sum_{k=1}^{M_i}p_k}{M_i}$, where $p_k$ is the coordinate of the nose tip of the $k$-th face. Furthermore, the relative distance $\delta_{k}$ of the $k$-th face is described as $\delta_{k}=\parallel p_k-c_g\parallel$, and $\delta_{k}$ is further normalised based on the mean relative distance. Based on $S_k$ and $\delta_{k}$, a global weight $w_k$ of $\mathbf{x}_k$ is obtained by as follows,
	\begin{equation}
	\label{eqn:globalweight}
	w_k=\parallel 1-\lambda\delta_{k}\parallel*S_{k},
	\end{equation}
	where $\lambda$ controls the effect of these weight factors on the global weight. In our method, we empirically set $\lambda$ as 0.1.
	
	For $\mathbf{x}_1,\ldots,\mathbf{x}_{M_i}$, their corresponding weights $w_1,\ldots,w_{M_i}$ are obtained by using Equation~\ref{eqn:globalweight}. According to the index of decreasing global weights, $\mathbf{x}_1,\ldots,\mathbf{x}_{M_i}$ are sorted. For two images $\mathbf{X}_i$ and $\mathbf{X}_j$, their face feature sets are sorted by using global weight sort method, namely $\widehat{\mathbf{X}}_i=\{\widehat{\mathbf{f}}_1,\ldots,\widehat{\mathbf{f}}_{M_i}\}$ and $\widehat{\mathbf{X}}_j=\{\widehat{\mathbf{g}}_1,\ldots,\widehat{\mathbf{g}}_{M_j}\}$, where $M_i$ and $M_j$ are the number of faces existed in $\mathbf{X}_i$ and $\mathbf{X}_j$, respectively. For the sake of simplicity, we remove~~$\widehat{}$~~out of $\widehat{\mathbf{X}}_i$ and $\widehat{\mathbf{X}}_j$ in the following discussion.
	
	\subsubsection{Similarity measurement function for two groups of faces}
	
	We aim to align $\mathbf{X}_i$ and $\mathbf{X}_j$ in various ways by distorting them. An alignment $\pi$ has length $P$ and $P<M_i+M_j-1$, since the two face sets have $M_i+M_j$ points and they are matched at one point of alignment path. An alignment $\pi$ is a pair of increasing integral vectors $(\pi_1,\pi_2)$ of length $p$ such that $1=\pi_1(1)\leq\ldots\pi_1(P)=M_i$ and $1=\pi_2(1)\leq\ldots\pi_2(P)=M_j$, with unitary increments and no simultaneous repetitions. Let $\mid\pi\mid$ denote the length of alignment $\pi$. The similarity measurement function in Equation~\ref{eqn:problem} for $\mathbf{X}_i$ and $\mathbf{X}_j$ can be defined by means of a local divergence $\phi$ that measures the discrepancy between any two points $\mathbf{f}_{\pi_{1}(p)}$ and $\mathbf{g}_{\pi_{2}(p)}$ as follows,
	
	\begin{equation}
	\label{eqn:cost}
	s({\mathbf{X}_i,\mathbf{X}_j})=\sum_{p}^{\mid\pi\mid}\phi(\mathbf{f}_{\pi_{1}(p)}, \mathbf{g}_{\pi_{2}(p)}).
	\end{equation}
	
	For resolving Equation~\ref{eqn:cost}, the global alignment is proposed to calculate the similarity of images, as it regards that the minimum value of alignments may be sensitive to peculiarities of the time series and uses all alignments weighted exponentially. It can be further defined as the sum of exponentiated and sign changed costs of the individual alignments like $k(\mathbf{X},\mathbf{Y})=\sum_{\pi\in A(m,n)}e^{(-s_{\mathbf{X},\mathbf{Y}}(\pi))}$. According to divergence $\phi$ and Equation~\ref{eqn:cost}, global alignment kernel is therefore formulated as follows,
	
	\begin{equation}
	\label{eqn:math1}
	k(\mathbf{X}_i,\mathbf{X}_j)=\sum_{\pi\in A(m,n)}\prod_{p}^{\mid\pi\mid}e^{-\phi(\mathbf{f}_{\pi_{1}(p)}, \mathbf{g}_{\pi_{2}(p)})},
	\end{equation}
	where $A(m,n)$ is the set of all alignments between two feature sets $\mathbf{X}_i$ and $\mathbf{X}_j$ of length $M_i$ and $M_j$.
	
	It has been argued by~\cite{GA} that $e^{-\phi}$ in Equation~\ref{eqn:math1} goes through the whole spectrum of the costs along all alignments. Additionally, it gives rise to a smoother measure than the minimum of the costs of some classical time-series alignment such as Dynamic Time Warping (\textbf{DTW}). Following the suggestion by~\cite{GA}, we use local kernel described as follows,
	\begin{equation}
	\label{eqn:KGA}
	k_{\text{GA}}(\mathbf{X}_i,\mathbf{X}_j)=\sum_{\pi\in A(m,n)}\prod_{p}^{\mid\pi\mid}e^{-\phi_{\sigma}},
	\end{equation}
	where $\phi_{\sigma}=\frac{1}{2\sigma^2}d(\mathbf{f}_{\pi_{1}(p)},\mathbf{g}_{\pi_{2}(p)})+\log(2-e^{-\frac{d(\mathbf{f}_{\pi_{1}(p)},\mathbf{g}_{\pi_{2}(p)})}{2\sigma^2}})$, $d$ is the distance function, and $\sigma$ is standard deviation.
	
	Based on Equation~\ref{eqn:KGA}, the basic form of SVR in Equation~\ref{eqn:SVR} is re-written as follows,
	\begin{equation}
	\label{eqn:SVRKGA}
	g(\mathbf{Y})=\sum_{i=1}^{N}\omega_il_iK_{\text{GA}}(\mathbf{X}_i,\mathbf{Y})+b,
	\end{equation}
	where $N$ is the number of images in the training set, $K_{\text{GA}}(\mathbf{X}_i,\mathbf{Y})$ is a GAK consisting of series of $k_{\text{GA}}(\mathbf{x},\mathbf{y})$.
	
	\subsection{Combined Global Alignment Kernels with application to SVR}
	\label{sec:CGAK}
	
	Aforementioned in Section~\ref{sec:facialFeature}, due to the challenging situations of group-level happiness intensity estimation, we exploited RVLBP and deep CNN features for facial images. In this section, we propose a simple but efficient way to adapt them for GAK for group-level happiness intensity estimation.
	
\begin{figure}[th!]
		\centering
		\includegraphics[width=0.9\linewidth]{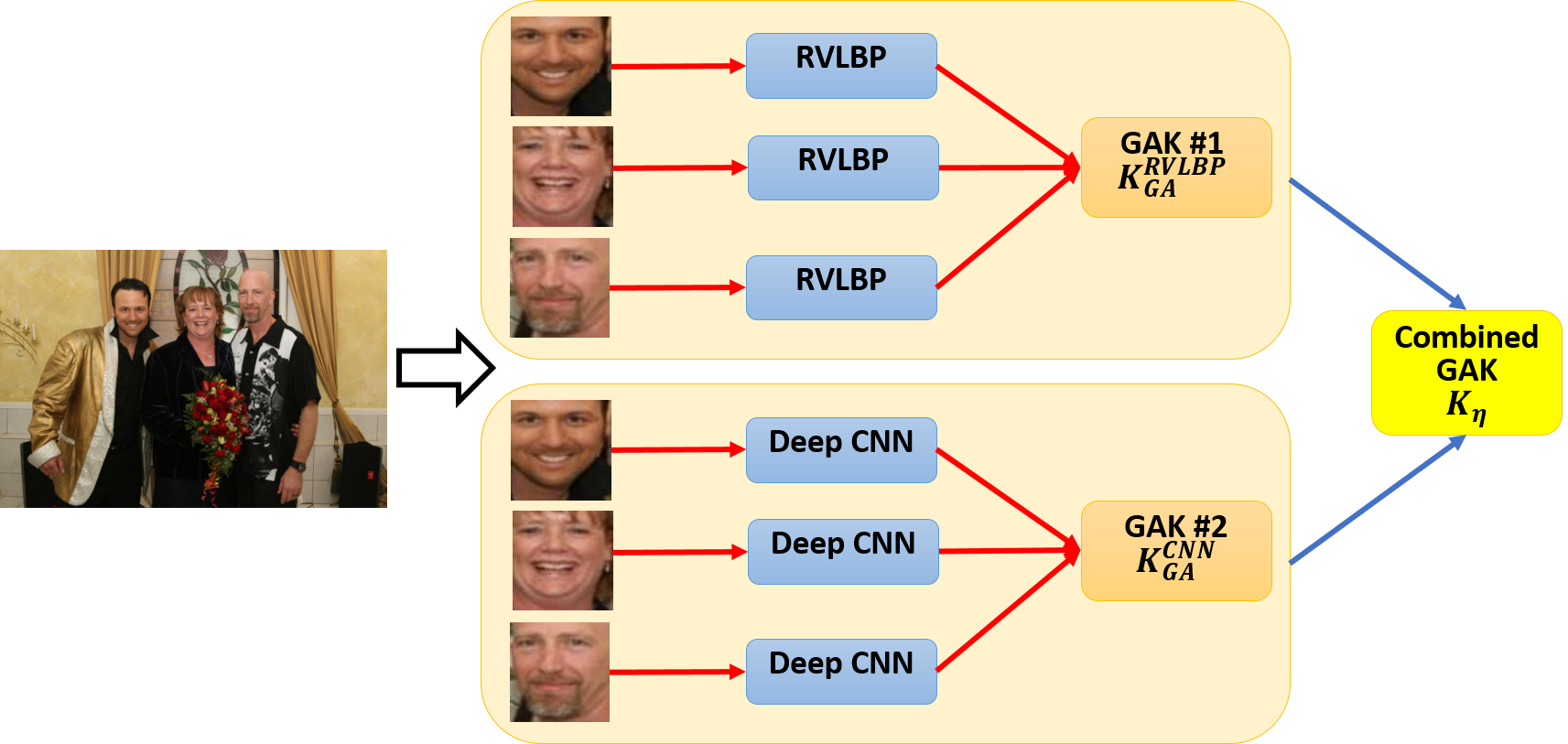}
		\caption{Illustration of SVR based on Combined Global Alignment Kernels (SVR-CGAK). For each face, RVLBP and deep CNN features are extracted. For each feature set, the specific global alignment kernel is generated. They are denoted as $K_{\text{GA}}^{\text{RVLBP}}$ and $K_{\text{GA}}^{\text{CNN}}$ for RVLBP and deep CNN features, respectively. Subsequently, the combination strategy is employed to fuse both kernels.}
		\label{fig:SVR-CGAK}
	\end{figure}

	Considering RVLBP and deep CNN features, a convenient way is to concatenate them into one feature vector $\mathbf{X}$ and then feed to Equation~\ref{eqn:SVRKGA}. However, the feature concatenation method fails to consider the complementary information between both features. Instead, we attempt to combine them in an alternative way. Amongst combining kernel methods, an simple but effective one, namely fixed rules use the combination function as a fixed function of the kernels, without any training. Once we calculate the combined kernel, we train a single kernel machine using this kernel~\cite{Gonen2011,Gonen2013}. For example, we can obtain a valid kernel by taking the summation or multiplication of two kernels. One advantage of fixed rules is that they do not require parametrisation such that they can obtain efficient computation. The summation rule is applied successfully in predicting protein-protein interactions~\cite{Ben2005}, computational biology~\cite{Pavlidis2001} and optical digit recognition~\cite{Moguerza2004} to combine two or more kernels obtained from different representations. Considering the efficiency of fixed rules, we propose a SVR with combined GAK, namely \textbf{SVR-CGAK}, in which combined GAK is achieved by a fixed rule combines two kernels. SVR-CGAK is briefly illustrated in Figure~\ref{fig:SVR-CGAK}.
	
	In our method, RVLBP and deep CNN features are considered to construct the similarity measurement function. In general, the squared Euclidean distance is typically used to define the local divergence in Equation~\ref{eqn:cost}~\cite{FGA}. Instead, for RVLBP, Chi-Square distance may better characterize the similarity than squared Euclidean distance. We choose the Chi-Square distance~\cite{McCune2002} defined as $d(\mathbf{f},\mathbf{g})_{\text{RVLBP}}=\sum_{k=1}^{n}\frac{(\mathbf{f}_k-\mathbf{g}_k)^2}{(\mathbf{f}_k+\mathbf{g}_k)}$ where $n$ is a number of bins, and the square Euclidean distance denoted as $d(\mathbf{f},\mathbf{g})_{\text{CNN}}=\parallel \mathbf{f}-\mathbf{g}\parallel^2$ for RVLBP and deep CNN features, respectively. Based on these designed kernel, we aim to combine both kernels by considering their complementary information. Among the fixed rules, three general strategies are commonly adopted in multiple kernel learning~\cite{Pavlidis2001,Moguerza2004,Ben2005}: summation, multiplication and weighted summation rules, which are described as followed,
	
	\begin{itemize}
		\item
		Summation strategy~\cite{Gustau2009}:
		\begin{equation}
		\label{eqn:sum}
		K_{\eta}=K_{\text{GA}}^{\text{RVLBP}}(\mathbf{X}_i,\mathbf{Y})+K_{\text{GA}}^{\text{CNN}}(\mathbf{X}_i,\mathbf{Y}).
		\end{equation}
		\item
		Multiplication strategy~\cite{Haussler1999}:
		\begin{equation}
		\label{eqn:multi}
		K_{\eta}=K_{\text{GA}}^{\text{RVLBP}}(\mathbf{X}_i,\mathbf{Y})\times K_{\text{GA}}^{\text{CNN}}(\mathbf{X}_i,\mathbf{Y}).
		\end{equation}
		\item
		Weighted summation strategy~\cite{Gonen2011}:
		\begin{equation}
		\label{eqn:MKL}
		K_{\eta}=\beta_{\text{RVLBP}}K_{\text{GA}}^{\text{RVLBP}}(\mathbf{X}_i,\mathbf{Y})+\beta_{\text{CNN}}K_{\text{GA}}^{\text{CNN}}(\mathbf{X}_i,\mathbf{Y}),
		\end{equation}
		where $\beta_{\text{RVLBP}}$ and $\beta_{\text{CNN}}$ are the weights for RVLBP and deep CNN features, respectively.
	\end{itemize}
	
	With the combination strategy, the SVR is given by
	\begin{equation}
	\label{eqn:SVRCGAK}
	g(\mathbf{Y})=\sum_{i=1}^{N}\omega_il_iK_{\eta}(\mathbf{X}_i,\mathbf{Y})+b.
	\end{equation}
	
	\begin{lemma}
		\label{lemma:pd}
		Let $A_{i}$ be positive definite matrix. If $\lambda_{i}>0$ is a real number, then $\lambda_{i} A_{i}$ is positive definite. The sum $\sum_{i}\lambda_iA_{i}$ and multiplication $\Pi_{i}A_{i}$ are positive definite.
	\end{lemma}
	
	Cuturi~et.al~\cite{GA} proved that $k_{\text{GA}}$ is positive definite. According to Lemma~\ref{lemma:pd} observed by~\cite{Horn1985}, it is ensured that the three aforementioned combined kernels are positive definite. Therefore, the SVR based on combined GAK is a convex optimization problem. It can be efficiently solved by quadratic programming algorithm.
	
	\subsubsection{Solution to weighted summation strategy}
	\label{sec:weight}
	Amongst three combined kernels, it is observed that the weighted summation strategy contains two weights for RVLBP and deep CNN features, respectively. It is seen that there is an important issue to obtain the optimal weights $\beta_{\text{RVLBP}}$ and $\beta_{\text{CNN}}$ for the combined kernel. For resolving this problem, we propose to use localized multiple kernel regression~\cite{Gonen2010} to obtain the optimal weights.
	
	According to~\cite{Gonen2010}, Equation~\ref{eqn:SVRCGAK} can be re-formulated including two kernels as followed,
	\begin{equation}
	\begin{split}
	\label{eqn:svr}
	&\min \frac{1}{2}(\|\omega_{\text{RVLBP}}\|^2+\|\omega_{\text{CNN}}\|^2)+C\sum_{i=1}^{N}(\xi_i^{+}+\xi_i^{-}) \\
	&\text{w.r.t.}~~ \omega_{\text{RVLBP}}, \omega_{\text{CNN}}, b, \xi_i^{+}, \xi_i^{-}, \beta_{\text{RVLBP}}, \beta_{\text{CNN}} \\
	&\text{s.t.}~~~ \epsilon+\xi_i^{+}\geq y_i, \epsilon+\xi_i^{-}\leq y_i, \\
	&~~~~~~~~\xi_i^{+}\geq 0, \xi_i^{-} \geq 0,  \forall i, \\
	\end{split}
	\end{equation}
	where $C$ is the regularization parameter, $\{\xi, \xi^{+}, \xi^{-}\}$ are slack variables, and $\epsilon$ is the tube width. Its dual formulation is easily obtained as:
	\begin{equation}
	\begin{split}
	\label{eqn:rlmkr_obj}
	&\max \sum_{i=1}^{N}l_i(\omega_i^{+}-\alpha_i^{-})+\epsilon\sum_{i=1}^{N}l_i(\alpha_i^{+}+\alpha_i^{-}) \\
	&~~~~~~~~~-\frac{1}{2}~(\alpha_i^{+}-\alpha_i^{-})(\alpha_j^{+}-\alpha_j^{-})K_{\eta}, \\
	&~\text{w.r.t.}~~ \alpha_i^{+}, \alpha_i^{-} \\
	&~\text{s.t.}~ \sum_{i=1}^{n}(\alpha_i^{+}-\alpha_i^{-})=0\\
	&~~~~~~~~C\geq \alpha_i^{+}\geq 0, C\geq \alpha_i^{-}\geq 0, \forall i \\
	\end{split}
	\end{equation}
	
	In~\cite{Gonen2010}, the parametric gating model was proposed to assign a weight to feature space as function of feature and a vector of gating model parameters. For weighted summation strategy, we choose the updated gating model as the weights for RVLBP and deep CNN features, respectively. For the gating model, we implement Softmax function which can be expressed as:
	\begin{equation}
	\label{eqn:beta}
	\beta=\frac{exp(\langle v_m,\mathbf{X}_r\rangle+v_{r0})}{\sum_{q=1}^{Q}exp(\langle v_q,\mathbf{X}_q\rangle+v_{q0})},
	\end{equation}
	where $q$ represents RVLBP or CNN, $Q=2$, $v_q,v_{q0}$ are the parameters of this gating model and the Softmax guarantees non-negativity, respectively.
	
	For obtaining the optimal $\beta_{\text{RVLBP}}$ and $\beta_{\text{CNN}}$, we can simply use the objective function of Equation~\ref{eqn:rlmkr_obj} to calculate the gradients of the primal objective with respect to the parameters $v_q$ and $v_{q0}$ until the objective function convergence. After updating these parameters $v_q$ and $v_{q0}$, we can easily obtain the optimal $\beta$ using Equation~\ref{eqn:beta}.
	
	\section{Experiments}
	\label{sec:experiment}
	
	This section will evaluate the influence of parameters to our proposed approaches by performing experiments on HAPPEI~\cite{Dhall2015} database. It will also show the benefits of SVR-CGAK over state-of-the-art approaches for estimating the happiness intensity of a group of people on both databases. Following the protocol in~\cite{Huang2015}, we implement a 4-fold-cross-validation in our experiments, where 1,500 images are used for training and 500 for testing, repeating four times. Mean Absolute Error (\textbf{MAE}) is used as the metric for estimating happiness intensity of images.
	
	In this section, we firstly evaluate the affect of global weight sort and two parameters ($\sigma$ and metric measurement) in Equations~\ref{eqn:KGA} and~\ref{eqn:cost}, respectively, and the combination strategy to GAK on HAPPEI database. Furthermore, we compare SVR-GAK with SVR based on the state-of-the-art DTW approaches and three sequential methods. Finally, we compare SVR-CGAK with the state-of-the-art approaches~\cite{Huang2015} on HAPPEI database. 
	
	\subsection{Parameter evaluation}
	
	\textbf{Evaluation of the `global weight sort'}: Global weight sort method is developed based on graph for providing a consistent data structure to global alignment kernel. A general way, namely, `holistic way', is to use all neighbouring faces to each face for obtaining graph. However, this way may not provide the relative position of each face in a group. As well, it may bring the noise caused by the isolated faces to the graph. To alleviate these disadvantages, we use minimal spanning tree algorithm~\cite{Prim1957} to obtain a fully connected graph that can provide the spatial correlation between faces and mostly reduce the influence of the isolated faces.
	
	We conduct an experiment for comparing `global weight sort' method based on minimal spanning tree algorithm with one based on holistic manner on HAPPEI database. We use Local Binary Pattern (LBP)~\cite{Ojala2002}, Local Phase Quantization (LPQ)~\cite{Ojansivu2008}, Completed LBP (CLBP)~\cite{Guo2010}, RVLBP~\cite{Huang2015} and $\text{FC6-L2}$ to evaluate the influence of two aforementioned ways to global weight sort. To make a fair comparison, $\sigma$ and distance function for SVR-GAK are set as 10 and Square-Euclidean distance, respectively. The comparative results on HAPPEI database are presented in Table~\ref{tab:GWS}. We conduct t-tests for the comparisons between global weight sort based minimal spanning tree algorithm and holistic method alone over all features. We obtain p=.0863 ($p>.05$), which indicates that compared with holistic method, the improvements that are achieved by the minimal spanning tree algorithm are not significant. However, comparing with the decreased MAE across each feature, we can see that the improvement is still competitive and promising. Minimal spanning tree algorithm improves the performance over `holistic way'. For `holistic way', its poor performance may be caused due to noise made by the isolated faces.
	
	\begin{table}[th!]
		\small
		\caption{Performance comparison of using minimal spanning tree algorithm and holistic manner for global weight sort, where Mean Absolute Error is used as performance metric. The best result is in bold. Minimal spanning tree algorithm has considerable improvement on all features over holistic method.}
		\label{tab:GWS}
		\begin{center}
			\begin{tabular}{|l|c|c|}
				\hline
				\multirow{2}{*}{\textbf{Features}} & \multicolumn{2}{c|}{\textbf{Global Weight Sort}}\\			
				\cline{2-3}\multicolumn{1}{|c|}{}&\multicolumn{1}{c|}{\thead{Minimal spanning \\ tree algorithm}}&\multicolumn{1}{c|}{\textbf{Holistic method}}\\
				\hline			
				LBP~\cite{Ojala2002} & 0.5690 & 0.5717\\
				LPQ~\cite{Ojansivu2008} & 0.5674 & 0.5681\\
				CLBP~\cite{Guo2010} & 0.5665 & 0.5668\\
				RVLBP~\cite{Huang2015} & 0.5420 & 0.5421\\
				FC6-L2 & \textbf{0.526} & 0.5277\\
				\hline			
			\end{tabular}
		\end{center}
	\end{table}
	
	To demonstrate the advantage of `global weight sort', we compare SVR-GAK `with global weight sort' with one `without global weight sort' on HAPPEI database. Specifically, `without global weight sort' means that we use face detector~\cite{Everingham2006} to automatically localize multiple faces and output its subsequent face results according to its search order. Table~\ref{tab:dataOrder} tabulates the comparative results of SVR-GAK `without global weight sort' and `with global weight sort' on  HAPPEI database. We conduct t-tests for the comparisons between SVR-GAK with and without global weight sort alone over all features. We obtain p=.0185 ($p<.05$), which indicates that compared with SVR-GAK without global weight sort, the improvements that are achieved by SVR-GAK with global weight sort are significant. Additionally, we take RVLBP feature for SVR-GAK for example as analysis. According to Table~\ref{tab:dataOrder}, SVR-GAK obtains promising the lower MAE of 0.542 using global weight sort scheme. The SVR-GAK obtains promising results using global weight sort scheme comparing with SVR-GAK `without global weight sort'. This may be explained by the following: (1) we used `global weight sort' scheme to extract the consistent structure of faces in image and (2) SVR-GAK obtain the well optimal path between faces in two images based on that structure. On the other hand, comparing SVR-GAK without global weight sort, SVR-GAK based on global weight sort scheme obtains considerable improvement over all features. As was mentioned in Section~\ref{sec:GWS}, global weight sort scheme sorts the face according to their importances in the image. Such way can efficiently provide the consistent graph structure when SVR-GAK computes the optimal path between faces in two images. Comparative results give us an evidence that the significant sorted related position of faces can affect the performance of SVR-GAK. They also demonstrates that consistent graph structure provide a positive support to SVR-GAK.
	
	\begin{table}[th!]
		\small
		\caption{Performance comparison of SVR-GAK `without global weight sort' and `with global weight sort' on HAPPEI database, where Mean Absolute Error is used as performance metric. The best result is highlighted with bold. Global weight sort scheme has considerable improvement over all features.}
		\label{tab:dataOrder}
		\begin{center}
			\begin{tabular}{|l|c|c|c|}
				\hline
				\multirow{2}{*}{\textbf{Features}} &  \multicolumn{2}{c|}{\textbf{Methods}}\\			
				\cline{2-3}\multicolumn{1}{|c|}{}&\multicolumn{1}{c|}{\thead{Without Global\\Weight Sort}}&\multicolumn{1}{c|}{\thead{With Global\\ Weight Sort}}\\
				\hline			
				LBP~\cite{Ojala2002} & 0.5992 & 0.5690\\
				LPQ~\cite{Ojansivu2008} & 0.5810 & 0.5674\\
				CLBP~\cite{Guo2010} & 0.5689 & 0.5665\\
				RVLBP~\cite{Huang2015} & 0.5671 & 0.5420\\
				FC6-L2 & 0.5490 & \textbf{0.526}\\
				\hline			
			\end{tabular}
		\end{center}
	\end{table}
	
\begin{figure}[th!]
		\centering
		\includegraphics[width=0.9\linewidth]{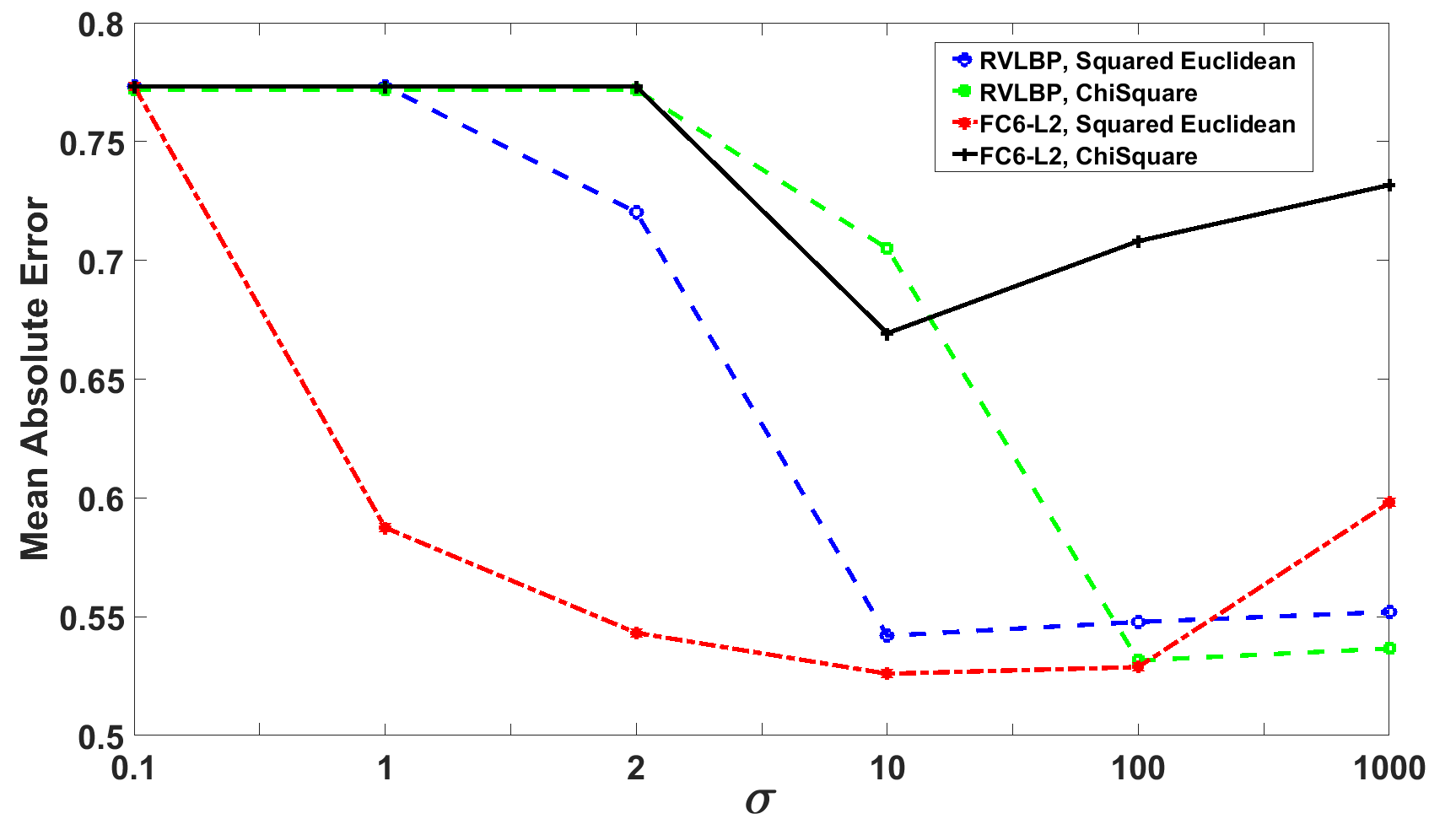}
		\caption{Influence of $\sigma$ and metric measurements to the performance of SVR-GAK on RVLBP and $\text{FC6-L2}$.}
		\label{fig:sigma}
	\end{figure}

	\textbf{The influence of $\sigma$ and metric measurement}: We evaluate the affects of $\sigma$ in $\{0.1, 1, 2, 10, 100, 1000\}$ for Equation~\ref{eqn:KGA} and two metric measurements (`Squared Euclidean' and `ChiSquare') for Equation~\ref{eqn:cost}. We choose one of RVLBP and $\text{FC6-L2}$ as feature descriptor for SVR-GAK . Due to experimental setup, it thus derives four different cases: (1) `RVLBP, Squared Euclidean', (2) `RVLBP, ChiSquare', (3) `FC6-L2, Squared Euclidean', and (4) `FC6-L2, ChiSquare'. The parameter evaluation across $\sigma$ and metric measurement on HAPPEI database is illustrated in Figure~\ref{fig:sigma}. 
	
	As we can see, for four cases, the performance is sensitive to varied $\sigma$. Especially, the performance is obviously improved with the increasing $\sigma$. A small $\sigma$ will make the GA kernel function in Equation~\ref{eqn:KGA} with a large variance. It also implies the support vector obtained in small $\sigma$ will have influence on regression. The increasing $\sigma$ will lead to high bias and low variance models. It implies the support vector does not have wide-spread influence. For the most of four cases, the promising $\sigma$ is dropped into the range from 2 to 100. It is seen that when $\sigma$ reaches 10, SVR-GAK performs considerably better than other $\sigma$ values. It may be explained that SVR-GAK reaches the good balance between bias and variance.
	
	Additionally, it is seen that ChiSquare metric works worse than Squared Euclidean for $\text{FC6-L2}$. It implies that ChiSquare metric is not suitable to $\text{FC6-L2}$. As observed from Figure~\ref{fig:sigma}, we choose ChiSquare and Squared Euclidean for RVLBP and $\text{FC6-L2}$, respectively. Additionally, $\sigma$ is set as 100 and 10 for RVLBP and $\text{FC6-L2}$, respectively.
	
	\textbf{Evaluation of the combination strategy}: Using the well-designed $\sigma$ and metric measurement, we evaluate the influence of `summation', `multiplication' and `weighted summation' combination strategies, which are formulated in Equations~\ref{eqn:sum},~\ref{eqn:multi} and~\ref{eqn:MKL}, respectively for SVR-CGAK. Table~\ref{tab:combinedGAK} tabulates the results of SVR-CGAK using summation~\cite{Gustau2009}, multiplication~\cite{Haussler1999} and weighted summation~\cite{Gonen2011} strategies in term of Mean Absolute Error (MAE) on HAPPEI database. According to Table~\ref{tab:combinedGAK}, SVR-CGAK achieves the best performance by using `weighted summation' rule, followed by `summation', `multiplication'. This result demonstrates that the `summation' rule outperforms the `multiplication' rule, while learning optimal weights in `weighted summation' can significantly boost the performance.
	
	According to Table~\ref{tab:combinedGAK}, we further discuss the comparison among three rules. As we see, the `summation' rule considerably improves the performance of 0.0069 compared with `multiplication'. It is explained by two things: (1)  `multiplication' rule has the disadvantage that it stores statistically significantly more support vectors than summation rule~\cite{Gonen2011}; (2) `multiplication' rule has very small kernel values at the off-diagonal entries of the combined kernel matrix. Comparing with `summation' rule, the performance of SVR-CGAK can be improved to 0.5082 in the term of MAE when we adopt `weighted summation' strategy, as the learned weight by MKL provides an effective way for improving the ability of SVR-CGAK based on summation strategy. It is seen that `weighted summation' strategy estimates the optimal weights of the basis kernels through optimizing a target function. The target function is a parametric function of the kernel weights that reaches its extremum on the best set of kernel weights.
	
	\begin{table}[th!]
		\small
		\caption{Performance comparison of SVR-CGAK using `summation' (Equation~\ref{eqn:sum}), `multiplication' (Equation~\ref{eqn:multi}) and `weighted summation' (Equation~\ref{eqn:MKL}) strategies, respectively. Bold number symbolizes best result. Mean Absolute Error is used as performance metric. Amongst, `weighted summation' achieves the best, followed by `summation' and `multiplication'. The learned weight in Section~\ref{sec:weight} provides an efficient way for summation strategy.}
		\label{tab:combinedGAK}
		\begin{center}
			\begin{tabular}{|l|c|c|c|}
				\hline			
				\multirow{2}{*}{\textbf{}} & \multicolumn{3}{c|}{\textbf{Strategy}}\\			
				\cline{2-4}\multicolumn{1}{|c}{}&\multicolumn{1}{|c}{Summation}&\multicolumn{1}{|c}{Multiplication}&\multicolumn{1}{|c|}{Weighted summation}\\
				\hline					
				MAE & 0.5144 & 0.5213 & \textbf{0.5082}\\			
				\hline			
			\end{tabular}
		\end{center}
	\end{table}
	
	\begin{figure}[th!]
		\centering
		\includegraphics[width=0.9\linewidth]{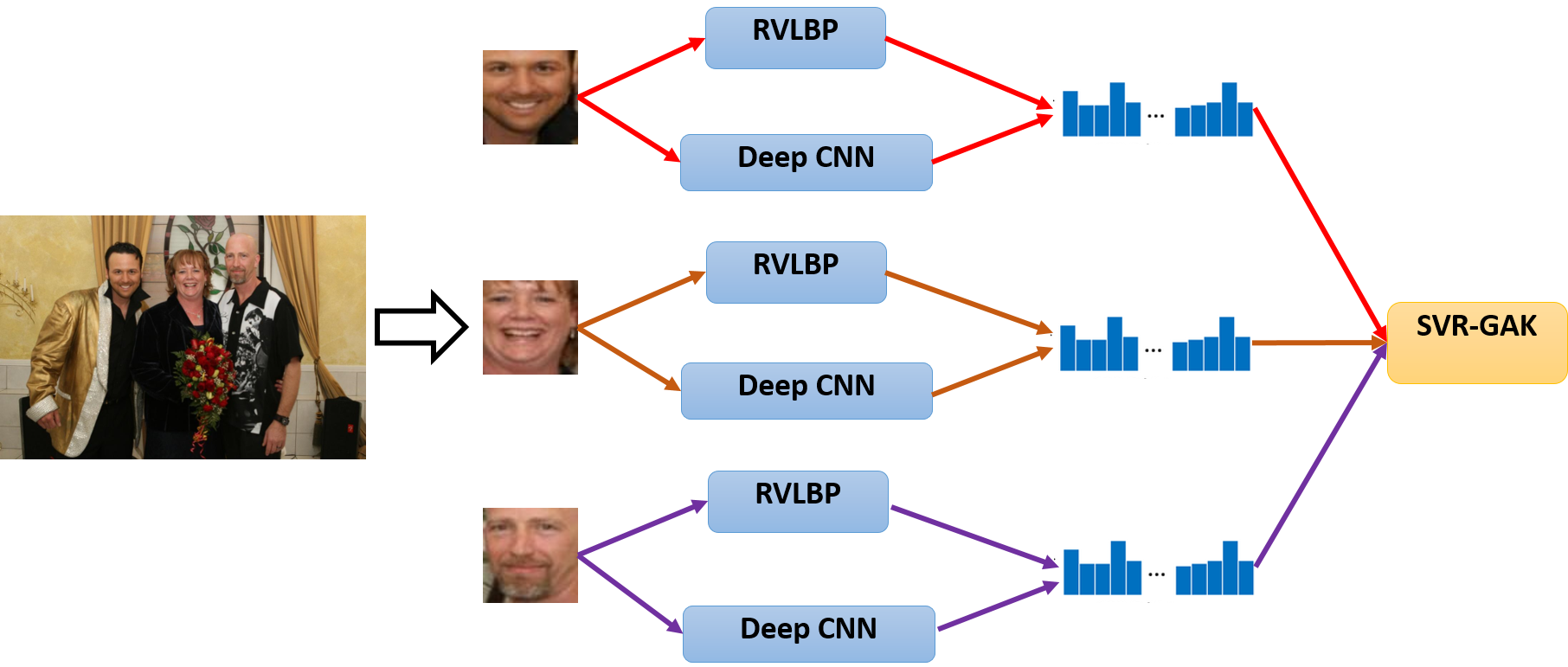}
		\caption{Illustration of SVR-GAK based on feature concatenation method (SVR-GAK-FC). For each face, RVLBP and deep CNN features are extracted. Then both features are concatenated into one feature vector. After that, this feature vector is fed into global alignment kernel (GAK).}
		\label{fig:SVR-GAK-FC}
	\end{figure}
	
	\begin{figure}[th!]
		\centering
		\includegraphics[width=0.9\linewidth]{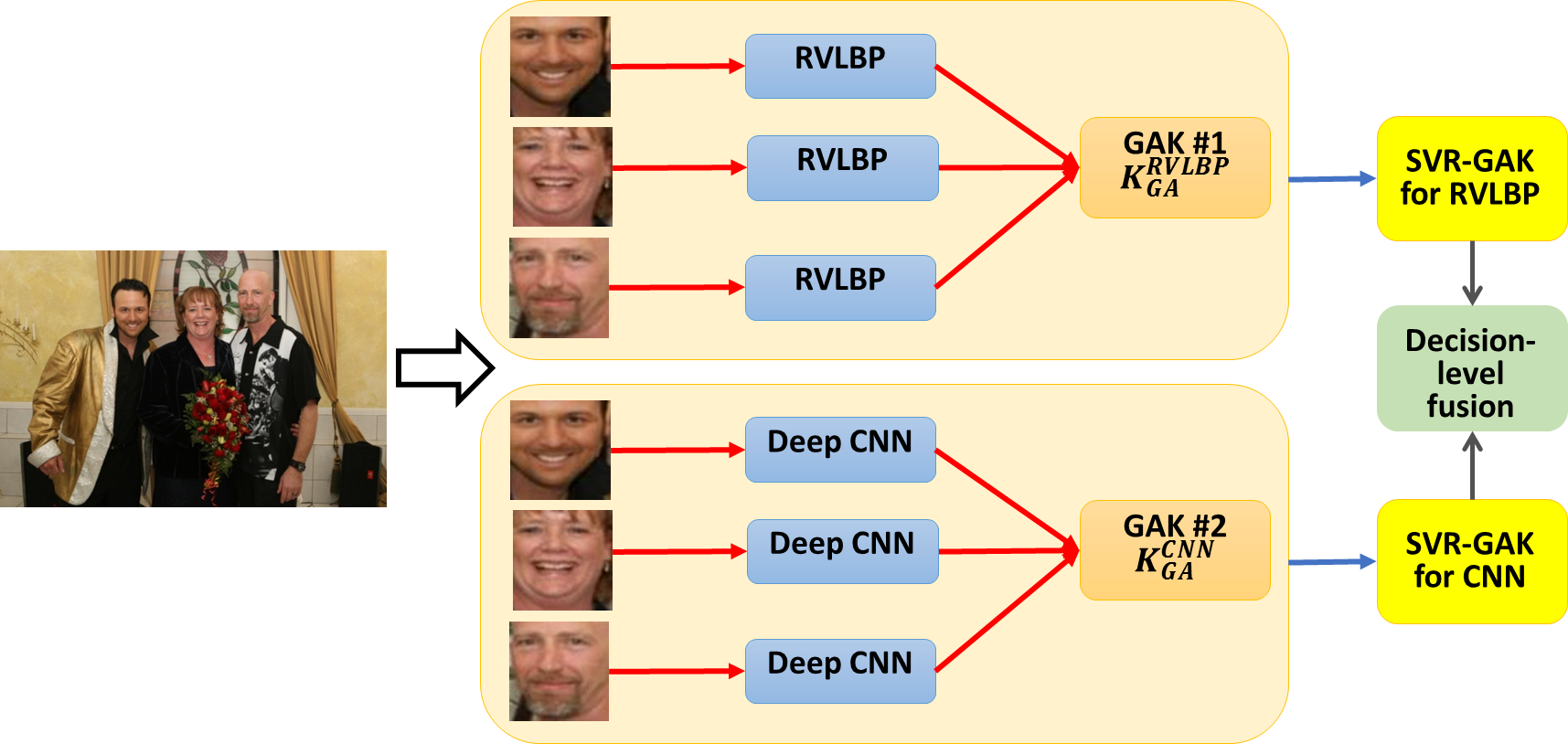}
		\caption{Illustration of decision-level fusion on SVR-GAK. Firstly, faces are detected and aligned from an image. Secondly, RVLBP and deep CNN features are employed to extract their corresponding features, respectively. For each image, we construct two SVR-GAK classifiers based on RVLBP and deep CNN features, respectively. In the final level, we use mean-rule to average the happiness intensity of an image.}
		\label{fig:SVR-GAK-Decision}
	\end{figure}
	
	\begin{table*}[t!]
		\small
		\caption{Performance and computation comparison of SVR based on the three DTW kernels and GAK on HAPPEI database, where TSK means time series kernel and the number inside the bracket represents average computation time (millisecond) on calculating kernel of pair-wise images. Mean Absolute Error (MAE) is employed as performance metric. The best result is in bold. SVR-GAK outperforms SVR based on all DTW kernels. Although computing GAK needs more time than calculating DTW kernels, it has still competitive in efficient computation.}
		\label{tab:compGA}
		\begin{center}
			\begin{tabular}{|l|c|c|c|c|c|c|}
				\hline			
				\multirow{2}{*}{\textbf{TSK}} & \multicolumn{6}{c|}{\textbf{Features}}\\			
				\cline{2-7}\multicolumn{1}{|c}{}&\multicolumn{1}{|c}{LBP~\cite{Ojala2002}}&\multicolumn{1}{|c}{LPQ~\cite{Ojansivu2008}}&\multicolumn{1}{|c}{CLBP~\cite{Guo2010}}&\multicolumn{1}{|c|}{RVLBP~\cite{Huang2015}} & \multicolumn{1}{|c|}{FC6-L2} & \multicolumn{1}{|c|}{RVLBP+FC6-L2}\\			
				\hline			
				SVR-DTW & \makecell{0.7091 \\ (0.0806)} & \makecell{0.7375 \\ (0.4160)} & \makecell{0.7123 \\ (0.1856)} & \makecell{0.7104 \\ (2.6653)} & \makecell{  0.7101 \\ (0.0669)} & \makecell{ 0.7112 \\ (2.8252)}\\
				SVR-NDTW & \makecell{0.6507 \\ (0.0838)} & \makecell{0.6795 \\ (0.3967)}& \makecell{0.7189 \\ (0.1812)} & \makecell{0.6328 \\ (2.6051) } & \makecell{  0.6302 \\ (0.0689)} & \makecell{ 0.6001 \\ (2.7821)}\\
				SVR-GDTW & \makecell{0.6485 \\ (0.0850)} & \makecell{0.8378 \\ (0.4218)} & \makecell{0.6343 \\ (0.1891)}  & \makecell{0.6219 \\ (2.619)} & \makecell{  0.6297 \\ (0.0653)} & \makecell{ 0.6008 \\ (2.7998)}\\
				SVR-GAK & \makecell{0.5690 \\ (0.104)} & \makecell{0.5674 \\ (0.4792)} & \makecell{0.5665 \\ (0.2176)} &  \makecell{0.5420 \\ (3.0024)} & \makecell{ 0.5260 \\ (0.0702)} & \makecell{ \textbf{0.5220} \\ (3.2712)}\\
				\hline			
			\end{tabular}
		\end{center}
	\end{table*}
	
	\textbf{Performance comparison of SVR-CGAK, SVR-GAK-FC and SVR-GAK-DC}: To justify that the proposed SVR-CGAK works consistently well, we further exploit SVR-GAK based on feature concatenation method, namely `SVR-GAK-FC' and SVR-GAK based on decision-level, namely, `SVR-GAK-DC', as a comparison. 
	
	`SVR-GAK-FC' approach is designed by two followed steps: (1) we concatenate RVLBP and deep CNN features of one face into one feature vector, (2) we feed these features into one GAK. The procedure is illustrated in Figure~\ref{fig:SVR-GAK-FC}. In the experimental setup, we set $\sigma$ as 10 and two distance metrics including Squared Euclidean and ChiSquare for SVR-GAK, respectively. The MAE for SVR-GAK-FC are 0.5220 and 0.7508 for Squared Euclidean and ChiSquare, respectively.
	
	We further design the architecture of `SVR-GAK-DC', as depicted in Figure~\ref{fig:SVR-GAK-Decision}. Firstly, RVLBP and deep CNN are implemented to extract the corresponding features of faces, namely, $\mathbf{g}_{RVLBP}$ and $\mathbf{g}_{CNN}$, respectively. Subsequently, two global alignment kernels are constructed for RVLBP and deep CNN features, denoted as $\mathbf{K}_{GA}^{RVLBP}$ and $\mathbf{K}_{GA}^{CNN}$, respectively. And they will lead to two SVR-GAK classifiers based on $\mathbf{K}_{GA}^{RVLBP}$ and $\mathbf{K}_{GA}^{CNN}$, respectively. In the final level, three fusion strategies (mean, production and weight mean) are considered to fuse the output of SVR-GAK for RVLBP and SVR-GAK for CNN in the SVR-GAK-DC.
	
	For decision-level, $\sigma$ was set as 100 and 10 for RVLBP and deep CNN, respectively. ChiSquare is used for RVLBP, while Square Euclidean for deep CNN. We obtain the results of SVR-GAK-DC based on mean, production and weight mean decision-level rules in term of the Mean Absolute Error (MAE), which are 0.5315, 4.6067 and 0.5203, respectively. It is seen that production rule cannot work well for fusing happiness intensity. Amongst all decision rules, weight mean rule achieves the best result (0.5203). It is interesting to see that the weight learned by localized multiple kernel regression can promisingly balance the importance of two features.
	
	Finally, we compare three combination strategies with feature concatenation (`SVR-GAK-FC') and~ decision-level (`SVR-GAK-DC'). For SVR-GAK-FC, it obtains the result of 0.5220 in term of MAE using weight mean decision-level rules. It is seen that SVR-CGAK based on one of three combination strategies outperforms SVR-GAK-FC. It may be explained by that (1) feature concatenation may result in a feature vector with very large dimensionality leading to the `curse of dimensionality' problem and (2) the concatenated features may be incompatible to distance metric.  We can also seen that three combination strategies can provide more efficient dimensionality reduction than feature concatenation method. On the other hand, we can see that SVR-CGAK based on `summation' (0.5144) or `weighted summation' (0.5082) strategy achieves better performance than SVR-GAK-DC (0.5203). For SVR-GAK-DC, it may be caused by the assumption of feature distribution independent for classifier fusion. Through these comparisons, it is seen that SVR-CGAK on three combination strategies overcomes the SVR-GAK-FC and SVR-GAK-DC.

	\subsection{Comparison with DTW}
	
	We compare SVR-GAK with SVR based on DTW and extensions of DTW~\cite{DTW3}, including DTW~\cite{DTW1}, Gaussian DTW (GDTW)~\cite{DTW4} and negated DTW (NDTW)~\cite{DTW3}. In the comparison, we re-organize all faces by using global weight sort. We conduct the comparison based on LBP, LPQ, CLBP, RVLBP, FC6-L2 and feature concentration of RVLBP and FC6-L2 (RVLBP+FC6-L2) for DTW and GAK, respectively. For GAK, we set $\sigma$ as 10 and Square Euclidean distance for all features. The comparative experiments were carried out on Matlab R2016b 64 bits with 3.1GHz Intel i5-2400 processor and 16Gb of RAM.
	
	Table~\ref{tab:compGA} reports the comparison of SVR based on three various DTW kernels and GAK in terms of MAE and computational time on HAPPEI database, where the number inside the bracket represents computation time on calculating kernel of pair-wise images. According to Table~\ref{tab:compGA}, we conduct t-test for the comparisons between SVR-GAK and SVR-DTW, SVR-NDTW, SVR-GDTW. We obtain p=5.06e-06 ($p<.05$), p=2.5e-04 ($p<.05$) and p=.0161 ($p<.05$), respectively. These results indicate that SVR-GAK achieves substantial improvement compared with SVR based on all DTW kernels. For example, when LBP is seen as input to kernels, SVR-DTW only obtains the MAE of 0.7091, which is worse than SVR-GAK. It is explained by that the gram matrix by GAK is positive definite but DTW is not. This positive definite condition may guarantee that SVR is not NP problem. Additionally, SVR-GAK achieves the best performance of 0.5220 in the term of MAE, when RVLBP+FC6-L2 is feed into GAK. It shows that feature descriptor has influence on the performance of SVR-GAK. It also gives us a tip that the appropriate feature descriptor will provide the similarity measurement more accurate. Furthermore, it is seen that GAK costs more than 0.4 ms comparing with variance of DTW in computing kernel function between two images. But GAK has still competitive in efficient computation.
	
	\subsection{Comparison with sequential methods}
	
	\begin{table*}[th!]
		\small
		\caption{Performance comparison of SVR-GAK with HMM~\cite{HMM2014}, CCRF~\cite{Imbrsaite2013} and RNN-LSTM~\cite{RNN2014}, where we reimplemented the code of HMM and CCRF for our comparision, while for RNN-LSTM we realize it. Mean Absolute Error is used as performance metric. The bold number means the best performance. In most of cases, SVR-GAK performs promising results.}
		\label{tab:compSeq}
		\begin{center}
			\begin{tabular}{|l|c|c|c|c|c|c|}
				\hline			
				\multirow{2}{*}{\textbf{Methods}} & \multicolumn{6}{c|}{\textbf{Features}}\\			
				\cline{2-7}\multicolumn{1}{|c}{}&\multicolumn{1}{|c}{LBP}&\multicolumn{1}{|c}{LPQ}&\multicolumn{1}{|c}{CLBP}&\multicolumn{1}{|c}{RVLBP}&\multicolumn{1}{|c}{FC6-L2}&\multicolumn{1}{|c|}{RVLBP+FC6-L2}\\			
				\hline		
				HMM & 0.6308 & 0.6198 & 0.6024 & 0.5881 & 0.5656 & 0.56 \\
				CCRF & 0.5886 & 0.5953 & 0.5734 & 0.526 & 0.5577 & 0.536 \\
				RNN-LSTM & 0.5864 & 0.5732 & 0.5548 & 0.5509 & 05266 & 0.5315\\
				SVR-GAK & 0.569 & 0.5674 & 0.5665 & 0.5420 & 0.5260 &  \textbf{0.5220} \\
				\hline			
			\end{tabular}
		\end{center}
	\end{table*}
	
	Sequential methods such as Hidden Markov Model (HMM) have been developed to resolve different length problem of face sequence in emotion recognition. In order to show the efficiency of SVR-GAK, we compare our proposed methods with HMM~\cite{HMM2014}, Continuous CRF (CCRF)~\cite{Imbrsaite2013} and RNN with long short term memory (RNN-LSTM)~\cite{RNN2014}. In the implementation, we consider LBP, LPQ, CLBP, RVLBP, FC6-L2 and RVLBP+FC6-L2 for HMM, CCRF and RNN-LSTM, respectively. The comparative results on HAPPEI database are described in Table~\ref{tab:compSeq}. 
	
	According to Table~\ref{tab:compSeq}, it is observed that HMM that achieves 0.56 in the term of MAE has the worst performance over all feature descriptors amongst all comparative methods. CCRF outperforms HMM, because CCRF is suitable to model the relationship between faces and intensity~\cite{Imbrsaite2013}. Furthermore, the RNN-LSTM obtains slightly better performance than HMM and CCRF. Furthermore, we conduct t-tests for the performance comparison between SVR-GAK with HMM, CCRF and RNN-LSTM. We obtain p=9.8e-05 ($p<.05$), p=.1032 ($p>.05$) and p=.2632 ($p>.05$), respectively. These results further indicate that SVR-GAK achieves significant improvement compared with HMM, but the improvements that are achieved by SVR-GAK are not significant compared with CCRF and RNN. However, these results show our method is still competitive to CCRF and RNN-LSTM over the all feature descriptors in estimating happiness intensity of a group of people. It gives us a tip that the deep net may be a potential explored method for group-level happiness intensity estimation. Overall, comparing with HMM, CCRF and RNN-LSTM, SVR-GAK has considerably competitive performance, as it borrows the superiority of SVR and good similarity measurement of GAK.
	
	\subsection{Comparison with state-of-the-art methods}
	
	Finally, we compare SVR-CGAK with several GEMs~\cite{Huang2015} including weighted GEM ($\text{GEM}_\text{weighted}$), LDA based GEM ($\text{GEM}_{\text{LDA}}$) and CCRF based GEM ($\text{GEM}_{\text{CCRF}}$) and information aggregation on face (INFA)~\cite{Huang2016} for group-level happiness intensity estimation as described as followed:
	
	(1) $\text{GEM}_\text{weighted}$: In~\cite{Huang2015}, the global and local context were extracted to formulate the relative weight $w_i$ for each face $g_i$ in an image. This relative weight $w_i$ is used to define a new group expression model like $\frac{\sum_{i=1}^{N}w_iI_{i}}{N}$, where $I_{i}$ is the happiness intensity of face $g_i$ predicted by using Kernel Partial Least Square regression and $N$ is the number of faces in an image.
	
	(2) $\text{GEM}_{\text{LDA}}$: For estimating happiness intensity, group expression model was proposed based on topic modelling and manually defined attributes for combining the global and local attributes.
	
	(3) $\text{GEM}_{\text{CCRF}}$: In $\text{GEM}_{\text{CCRF}}$, they firstly extracted RVLBP features from the detected faces of images. Furthermore, continuous conditional random fields (CCRF) was exploited to model the content information information of faces as well as the relation information between faces.
	
	(4) INFA: In~\cite{Huang2016}, they proposed an information aggregation based on an improved fisher vector to encode the facial regions from an image into a compact feature for an image. Specifically, they firstly divided facial images into several blocks and extracted their corresponding RVLBP features. Furthermore, a visual vocabulary is obtained based on lots of facial regions by using Gaussian Mixture Models (GMM). For each image, assume $N$ faces are detected, based on the visual vocabulary, the feature is obtained by the stacking the differences between the regional features and each of the GMM centers.
	
	The comparative results of the state-of-the-art algorithms, SVR-GAK (based on FC6-L2) and SVR-CGAK on HAPPEI database are shown in Table~\ref{tab:finalResult}, where their results are extracted from the recent works~\cite{Huang2015,Huang2016}. It is noted that the results are directly comparable due to the same experimental setups, preprocessing methods etc. We conduct t-tests for the performance comparison between SVR-GAK with the state-of-the-art methods on HAPPEI database. In addition, we perform t-tests to compare with SVR-CGAK with the state-of-the-art methods. We obtain p=.2966 ($p>.05$) and p=.3910 ($p>.05$), respectively. These results indicate that SVR-CGAK achieves comparative but not significant improvement comparing with $\text{GEM}_{\text{CCRF}}$, promisingly decreased by 0.021. It also obtains the lower MAE of 0.5082 than INFA. From intensive comparisons on HAPPEI database, SVR-CGAK achieves a considerable performance for group-level happiness intensity estimation.
	
	\begin{table}[th!]
		\small
		\caption{Mean Absolute Error (MAE) of the state-of-the-art algorithms and our proposed methods on HAPPEI database, where results of compared algorithms are directly from ~\cite{Huang2015,Huang2016}. The bold number is the best performance. INFA~\cite{Huang2016} has the lowest MAE amongst the state-of-the-art methods. Comparing with INFA, SVR-CGAK considerably improved the performance by the MAE of 0.0105.}
		\label{tab:finalResult}
		\begin{center}
			\begin{tabular}{|l|c|c|c|}
				\hline
				\textbf{Methods} & $\text{GEM}_\text{weighted}$ & $\text{GEM}_{\text{LDA}}$ & $\text{GEM}_{\text{CCRF}}$ \\
				\hline 
				MAE & 0.5469 & 0.5407 & 0.5292 \\
				\hline \hline
				\textbf{Methods} & INFA~\cite{Huang2016} & SVR-GAK & SVR-CGAK \\
				\hline
				MAE & 0.5187 & 0.526 & \textbf{0.5082}\\		 
				\hline			
			\end{tabular}
		\end{center}
	\end{table}
	
	\section{Conclusion}
	\label{sec:conclusion}
	
	To advance the research in affective computing, it is important to understand the affect exhibited by a group of people in images. In this paper, a simple but efficient method (SVR-CGAK) has been proposed to analyse the perceived affect of a group of people in an image. Firstly, we exploit RVLBP and deep CNN feature to characterize the facial images. Furthermore, we propose to use global alignment kernel (GAK) as a novel metric based on the consistent data structure to explicitly measure the similarity of two images. Finally, three combination strategies are proposed to fuse two GAKs based on deep CNN feature and RVLBP.
	
	We have conducted extensive experiments on the HAPPEI database for evaluating SVR-GAK and SVR-CGAK. Firstly, RVLBP and deep CNN features perform very well with SVR-GAK comparing with LBP, LPQ and CLBP. Secondly, the global weight sort approach provides a new consistent structure to SVR-GAK and promisingly increases the performance of SVR-GAK. More importantly, our results show feature-level fusion strategies are helpful to SVR-CGAK for predicting the perceived group mood more accurately. As well, computational efficiency comparison indicates SVR-CGAK has still competitive efficient computation. It is seen that SVR-CGAK is an efficient and effective way to estimate the happiness intensity of a group of people. In our future work, we will consider automatically selection of the optimal $\sigma$ for SVR-CGAK and extend the method to different group-level emotions.
	
	\bibliographystyle{IEEEtran}
	\bibliography{IEEEabrv,egbib}
	
\end{document}